\pdfoutput=1

\documentclass[11pt]{article}

\usepackage[]{emnlp2021}

\usepackage{times}
\usepackage{latexsym}

\usepackage{amsmath}
\usepackage{amsfonts}
\usepackage{graphicx}
\usepackage{enumitem}
\setlist{noitemsep}
\usepackage{booktabs}

\usepackage[T1]{fontenc}

\usepackage[utf8]{inputenc}

\usepackage{microtype}

\title{Lightweight Decoding Strategies for Increasing Specificity}

\author{Katy Ilonka Gero \\
  Columbia University\\
  \texttt{katy@cs.columbia.edu} \\ \And
  Chris Kedzie \\
  
  \texttt{kedzie@cs.columbiua.edu} \\
    \AND
  Savvas Petridis \\
  Columbia University \\
  \texttt{sdp2137@columbia.edu}  \\\And
  Lydia B. Chilton \\
  Columbia University \\
  \texttt{chilton@cs.columbia.edu}\\}

\begin{document}
\maketitle
\begin{abstract}
Language models are known to produce vague and generic outputs. 
We propose two unsupervised decoding strategies based on either word-frequency or point-wise mutual information to increase the specificity of 
any 
model that outputs a probability distribution over its vocabulary at generation time. 
We test the strategies in a prompt completion task; 
with human evaluations, we find that both strategies increase the specificity of 
outputs with only modest decreases in sensibility.
We also briefly present a summarization use case, where these strategies can produce more specific summaries.
\end{abstract}

\newcommand{\model}{p_\theta}
\newcommand{\token}{t}
\newcommand{\step}{i}
\newcommand{\context}{c}
\newcommand{\logit}{a}
\newcommand{\reals}{\mathbb{R}}
\newcommand{\modlogit}{b}
\newcommand{\tcount}{n}
\newcommand{\tcountmax}{n^*}
\newcommand{\vocab}{\mathcal{V}}
\newcommand{\clampmin}{w_0}
\newcommand{\clampmax}{w_1}
\newcommand{\scale}{k}
\newcommand{\PPMI}{\operatorname{PPMI}}
\newcommand{\PMI}{\operatorname{PMI}}
\newcommand{\occur}{n}
\newcommand{\corpussize}{s}
\newcommand{\LM}{LM}
\newcommand{\corpus}{p}

\section{Introduction}
Language models (LMs) are known to produce vague and generic outputs \citep{holtzman2019curious}. In domains like summarization \cite{fan2018controllable}, dialogue generation \cite{yao2016attentional}, and creative computing \cite{fan2018hierarchical},
outputs with higher specificity are often desired. While controlling the specificity
of model outputs has been explored previously, it is primarily 
approached as a supervised learning problem where access to large, in-domain
training sets are prerequisites for implementation. 

However, 
pre-trained LMs are increasingly of sufficient quality
such that only a text prompt is necessary to obtain 
a task specific language generator \cite{brown2020language}. 
It would be beneficial to control the specificity of these models' outputs in an unsupervised manner 
because re-training or fine-tuning such models are non-trivial or impossible tasks---for instance because the language model is too large, only accessible by an API,
or the generation task does not have training data.

To that end, we propose two unsupervised decoding strategies to increase the specificity of LMs that can work with any LM that outputs a probability distribution over its vocabulary at generation time. The first is based on word frequency and the second on positive point-wise mutual information (PPMI). We show in a prompt completion task that unsupervised reweighting strategies based on these quantities 
improves the specificity of generated outputs while only modestly affecting the sensibility according to human annotators. 

This paper has four main contributions:\footnote{Implementations and evaluations will be released after anonymous review.}

\begin{enumerate}
    \item We propose word frequency and PPMI based reweighting schemes of an LM's output probability distribution to increase specificity.
    \item We verify with human evaluations on a prompt completion task that these schemes improve specificity with only modest drops to
    sensibility. We find this holds both in deterministic 
    and stochastic generation settings.
    \item We verify with automatic measures that these schemes improve the diversity of outputs in deterministic generation settings.
    \item We show how these schemes can be used to control generated summaries of news articles.
\end{enumerate}

\section{Related Work}

Both word frequency and PPMI have been used in prior work to control
the specificity of generated outputs. \citet{yao2016attentional} 
train a dialogue generation model using a 
supervised learning objective and reinforcement learning to maximize the inverse document
frequency (IDF) of generated responses, which improves the quality of both generation and retrieval. Relatedly, \citet{ko2019domain} condition a decoder on a variety of measures (including word frequency) to improve specificity in dialogue generation and find that linguistically-driven measures generate the most informative and topical responses.
\citet{zhang2018learning}  train a neural dialogue generation model that takes as input the text of the previous utterance but also the normalized maximum inverse word frequency of the desired response, which significantly outperforms state-of-the-art models. 
\citet{takayama2020consistent} propose a similar approach,
using maximum PPMI between utterance and 
response as the specificity control mechanism.  

While all four works attempt to increase the specificity of generated outputs, they do so via training a 
language generator from scratch. Additionally, while \citet{yao2016attentional} only adds a loss function, \citet{ko2019domain}, \citet{zhang2018learning} and \citet{takayama2020consistent} add purpose built neural components to the decoder
to incorporate specificity controls, something that would be difficult to 
do with a large, pre-trained LM like GPT-3. By comparison,
our proposed unsupervised reweightings do not require retraining, fine-tuning, or  additional
decoder modifications and can work with any LM that produces a 
probability distribution
over next tokens. Being able to easily control specificity in such models as 
a light-weight post-processing step is crucial as most researchers do not
have the resources to train such models from scratch.






\section{Controlling Generation Specificity}

We present two ways to modify the probability distribution of a LM. 
The first relies on 
normalized inverse word frequencies (NIWF), which can be easily calculated using any desired corpus and doesn't depend on the prompt. The second relies on a calculation of positive point-wise mutual information (PPMI), which can be calculated using any desired corpus, but does rely on defining some context (likely a word or words from the prompt) for the calculation.

In either case, a corpus (which does not need to be the original training corpus) is used to modify the probability distribution coming out of the LM.
Both schemes modify the original distribution by adding a token specific term $\modlogit_\token \in \reals$
to the unnormalized log probability $\logit_{\token_\step}\in \reals$:
\begin{align}
\textit{(original model)} && \log\model(\token_\step) & \propto \logit_{\token_\step} \\
\textit{(reweighted model)} && \log\model(\token_\step) & \propto \logit_{\token_\step} + \modlogit_\token
\end{align}
where $\model(\token_\step)$ is the probability under the 
\LM~of generating token
$\token$ at step $\step$.\footnote{Typically $\model(\token_\step)$ will be conditional on the previously
generated tokens $\token_1,\ldots,\token_{\step-1}$ and optionally a context
$\context$ but we omit explicitly stating them here since they are not necessary to explain the reweighting schemes.}


\subsection{Normalized Inverse Word Frequency (NIWF)}

NIWF is often used to measure specificity \citep{li2015fast, ko2019domain}; here we use it to calculate a modified probability 
for each token $t_i$ in the model (at generation time).

Let $\tcount_\token$ be the count of token $\token$ in a corpus and let 
$\tcountmax = \max_{\token \in \vocab} \tcount_\token$ be the maximum count
occurring in the corpus. The NIWF reweighting  $\modlogit_\token$ of a token $\token$ is then calculated as:
\begin{equation}
\modlogit_\token = \min\Bigg(\max\left(\clampmin, \frac{\tcountmax}{ \scale \tcount_\token}\right), \clampmax \Bigg)
\end{equation}
where $\scale \in \reals$ is a scalar to adjust the range and $\clampmin,\clampmax \in \reals$ are lower and upper bounds respectively. We set $\scale=100$.
In practice, $\frac{\tcountmax}{\scale\tcount_\token}$ can vary quite widely.
To ensure the probability distribution of the model is not disturbed beyond recognition, we set $\clampmin=\exp\left(-5\right)$ and $\clampmax=1$ to bound $\modlogit_\token$ roughly between 0 and 1. The effect is that the rarest words receive an increase of at most 1 to the original 
$\logit_{\token_\step}$ term  while common words will receive almost no increase.

\subsection{Positive Point-wise Mutual Information (PPMI)}

PPMI is another measure often associated with term specificity \cite{takayama2020consistent} and measures the positive association between two events. 
This reweighting requires a context event between which to compute the PPMI of
the tokens from the model vocabulary. In our case let the context $\context \subset \vocab$ be a set of topically related words from a prompt text we would like the LM 
to complete (see section \ref{sec:implementation} for a concrete example).

We then define the modification term $\modlogit_\token$ to be
\begin{equation}
    \modlogit_\token = \max\Bigg(0, \;\log \frac{\corpus(\context, \token)}{\corpus(\context) \corpus(\token)}\Bigg)
\end{equation}
 
where $\corpus(\context)$, $\corpus(\token)$, and $\corpus(\context, \token)$ 
are the marginal probability of context words $\context$ occurring, the marginal probability 
of token $\token$ occurring, and the joint probability of context words $\context$
and token $\token$ co-occurring respectively. 

These probabilities are estimated
from a corpus of sentences with 
$\corpus(\context,\token) = \frac{\occur_{\context,\token}}{\occur_\corpussize}$, $\corpus(\context) = \frac{\occur_\context}{\occur_\corpussize}$, and 
$\corpus(\token) = \frac{\occur_\token}{\occur_\corpussize}$
where $\occur_\token$ is the number of sentences token $\token$ occurs in, $\occur_\context$ is the number of sentences the context words $\context$ occur in, $\occur_{\context,\token}$ is the number sentences where both a context word and 
$\token$ co-occur, and
$\occur_\corpussize$ is the size of the corpus in sentences.


\autoref{sec:logprobs} shows 
how these reweightings impact the log probabilities for a specific prompt.

\begin{table}
\centering
\begin{tabular}{ll}
\toprule
\textbf{condition} & \textbf{example outputs for prompt:} \\
& `Cryptography is used by' \\
\midrule
\multicolumn{2}{c}{beam search} \\
original & the world's largest companies\\
NIWF     & bitcoin miners    \\
PPMI     & Telegram apps   \\
\midrule
\multicolumn{2}{c}{top-$k$ sampling ($k$=50)} \\
original &  many applications\\
$\tau=1.7$  & many other crypto technologies\\
NIWF        & bitcoin wallet owners\\
PPMI        & privacy advocates\\
\bottomrule
\end{tabular}
\caption{Example outputs for the different conditions.}
\label{tab:ex-outputs}
\end{table}















\section{Experiment}

To test these methods, we use a science writing task where the model must produce a noun phrase about a technical topic. For instance, one prompt is ``Cryptography is used by''. This task requires the LM to say something \textit{sensible}, that makes sense given the topic, and \textit{specific}, that doesn't apply to just any topic. This is a difficult task for most pre-trained LMs which  tend to produce very vague outputs (e.g., completing the cryptography prompt with ``people'' or ``many''). 

We use five topics randomly sampled from Wikipedia's list of Computer Science topics:\footnote{\url{https://en.wikipedia.org/wiki/List_of_academic_fields\#Computer_sciences}}
 cryptography,
    human-computer interaction,
    support vector machines,
    databases,
    and 
    data structures.

For each topic we use four prompts to generate noun phrases: ``is used by'', ``is used in'', ``is studied by'', and ``is studied in". These prompts were selected for their ability to generate meaningful noun phrases about the topic. For each prompt we generate five output noun phrases. This set-up leads to 100 statements per condition ($5 \textrm{ topics} \times 4 \textrm{ prompts} \times 5 \textrm{ outputs}$) that can be scored for how \textit{sensible} and \textit{specific} the statement is.

We look at two generation paradigms: deterministic (beam search) and stochastic (top-$k$ sampling). For each paradigm, we have three conditions: original model (no reweighting), NIWF reweighting, and PPMI reweighting. Additionally, for top-$k$ sampling we also run the original model with a temperature parameter set to $\tau=1.7$ (selected such that the mean per word perplexity of outputs matches those from the NIWF reweighting scheme). \autoref{tab:ex-outputs} shows example outputs for each condition.

\subsection{Implementation Details}
\label{sec:implementation}

We use the Hugging Face implementation of GPT-2 (\texttt{gpt2-large})\footnote{\url{https://huggingface.co/gpt2-large}} as our pre-trained LM. 
To calculate the reweighting, we use a corpus of Vox news articles,\footnote{\url{https://data.world/elenadata/vox-articles}} which has over $30$ million tokens.

For the PPMI reweighting, we consider the context $\context$ to be the tokens making up the title of the computer science topic.\footnote{To reduce the sparsity of sentence level co-occurrence counts, for each topic
context $\context$ we also manually add morphologically related words 
(e.g. $\context = \{\textrm{cryptography}, \textrm{cryptographic}, \textrm{cryptographer}, \ldots \}$).}
For the top-$k$ sampling \citep{fan2018hierarchical} we set $k=50$.
To ensure outputs for each prompt are unique, we force the first token to be unique. For each prompt we generate the next 10 tokens, and use a parser to select the first noun phrase. See \autoref{sec:nounphrase} for details on noun phrase selection.

\begin{table*}[h]
\centering
\small
\begin{tabular}{p{0.98\textwidth}}
\toprule
\textbf{Baseline generation:} The Colonial Pipeline has restarted after a six-day shutdown. The pipeline's operators warned it will take several days for service to return to normal. The shutdown sparked panic-buying and hoarding that has overwhelmed gas stations in the Southeast. \\
\textbf{NIWF + market (economics):} The Colonial Pipeline has restarted after a six-day shutdown.
The pipeline was shut down after suffering a ransomware attack.
It provides nearly half the gasoline and \textit{diesel consumed by the East Coast}.
\textit{Oil industry executives} warned Wednesday that \textit{gas hoarding} is worsening the \textit{supply crunch}.\\
\textbf{PPMI + ransomware attack:} The Colonial Pipeline launched the restart of its operations Wednesday evening. The pipeline took itself offline Friday after suffering a \textit{ransomware} attack. The shutdown sparked panic-buying and hoarding that has overwhelmed gas stations in the Southeast. \\
\bottomrule
\end{tabular}
\caption{Results for generating summaries using specificity reweightings to encourage topical outputs. 
Italics indicate phrases related to the selected topic.}
\label{tab:summary-res}
\end{table*}

\subsection{Evaluation Methodology}

We have two human annotators score each statement for how sensible and specific it is.
We follow previous work on eliciting sensibility judgements from LM prompt completions \cite{li2016commonsense}
using a 0 -- 4 scale for sensibility, where 0 is ``Doesn't make sense`` and 4 is ``Generally true.'' We use a similar 0 -- 4 scale for specificity, where 0 is "Not sure if it applies" and 4 is "Applies to this topic in particular". We calculate a weighted Cohen's $\kappa$ to ensure adequate interannotation reliability, and average the annotators' scores if they differ. Each annotator is a PhD student in computer science with expert knowledge of the topics. For \textit{sensibility} we had an $\kappa = 0.35$ (fair agreement) and for \textit{specificity} we had a $\kappa = 0.53$ (good agreement).

We also calculate three diversity measures
following \citet{takayama2020consistent}. We report dist-1 and dist-2 (unigram and bigram uniqueness) and ent-2 (bigram-based entropy). See \autoref{sec:diversity} for details on the diversity measures.

\begin{table}
\centering
\begin{tabular}{llll}
\toprule
\textbf{scheme} & \textbf{sens} & \textbf{spec} & \textbf{dist1 / dist2 / ent2} \\
\hline
\multicolumn{4}{c}{beam search} \\
original    & \textbf{3.67}   & 1.27 & 0.32 / 0.54 / 4.17 \\
NIWF        & 3.13*   & 2.25* & \textbf{0.55} / \textbf{0.80} / \textbf{4.69}\\
PPMI        & 3.40*   & \textbf{2.39}* & 0.37 / 0.67 / 4.25\\
\midrule
\multicolumn{4}{c}{top-$k$ sampling ($k$=50)} \\
original    & 3.19   & 1.50 & 0.58 / 0.95 / 5.17\\
$\tau=1.7$       & 3.12  & 1.51 & 0.67 / \textbf{0.98} / \textbf{5.26} \\
NIWF        & \textbf{3.35}  & \textbf{2.27}*  & \textbf{0.70} / 0.97 / 4.98 \\
PPMI        & 3.26    & \textbf{2.27}* & 0.52 / 0.87 / 4.54\\
\bottomrule
\end{tabular}
\caption{Results of human \textit{sensibility} (sens) and \textit{specificity} (spec) evaluations, and diversity measures dist1, dist2, and ent2. Best (largest) result bolded. For sens and spec, * marks significant difference from original.}
\label{tab:main-res}
\end{table}

\subsection{Results}

The results for all measures can be seen in \autoref{tab:main-res}. We run significance tests (Mann-Whitney rank test for non-parametric data) on all conditions compared to the original model, and report significant results when $p < 0.001$. We found that the reweightings significantly  increase the specificity scores: in the deterministic case, NIWF increased absolute specificity by 0.98 and PPMI by 1.12; in the stochastic case, NIWF increased absolute specificity by 0.77 and PPMI by 0.77. 
Increasing the temperature of the distribution barely increased the specificity---by only 0.01 (not significant).

Additionally, in the stochastic setting, we find
that this increase in specificity actually slightly increases sensibility. The small decreases in bigram entropy (ent2) for stochastic NIWF and PPMI also suggest that the sampling distribution is more focused on topically specific words than the standard or increased temperature models.

In the deterministic setting, we saw modest, though significant, decreases in sensibility in the deterministic paradigm---NIWF decreased sensibility by 0.54 and PPMI by 0.27, a tradeoff found in prior work \citep{ko2019domain}. At the same time, the automatic metrics suggest that the reweightings, especially PPMI, improve the beam search diversity which is desired in many tasks  \citep{li-etal-2016-diversity}. 


\section{Use Case: Summarization}

To assess the generalizability of our specificity reweightings, we apply them to summarization. In \autoref{tab:summary-res} we compare baseline summarization to generating summaries with specificity reweightings. 
To compute the summaries, we use the Hugging Face implementation of (\texttt{pegasus}), fine-tuned on the CNN Dailymail dataset.\footnote{\url{https://huggingface.co/google/pegasus-cnn_dailymail}}
Each summary is generated from the same article on the Colonial Pipeline cyber-attack.\footnote{\url{https://www.cnn.com/2021/05/12/business/colonial-pipeline-restart/index.html}}
To calculate the NIWF reweighting, we calculate word counts from the Wikipedia page on ``Market (Economics)''. To calculate the PPMI reweighting, we use ``ransomware attack'' as our context and use the news article to obtain word and context co-occurrence counts. 
Both the NIWF corpus and PPMI context are hand-picked by the user. 

By incorporating a specificity reweighting, the summary is more focused on the selected topic. 
Compared to the standard summary, the NIWF summary includes more sentences pertaining to ``market'', including phrases like ``supply crunch'', ``gas hoarding'' and ``oil industry executives''. 
Similarly, the PPMI summary includes a specific sentence on the ransomware attack. 
With the reweightings, users can define a context to generate summaries focused on the topic of their interest.

\section{Conclusion}

We find that word frequency and PPMI based reweighting schemes increase language model specificity with modest to no decreases in sensibility. We demonstrate how these schemes can be used to control language model outputs in other tasks, like summarization.

\section{Broader Impacts Statement}

In this work we seek to improve the specificity of language model outputs by introducing lightweight decoding strategies. This work has both positive and negative impacts. The positive impacts include making the control of large, pre-trained language models more accessible to researchers and practitioners, as well as decreasing the compute costs (and therefore environmental and financial costs) of controlling large, pre-trained language models.

However, the use of large, pre-trained language models has been called into question given the gargantuan amounts of data they are trained on, which re-enforce hegemonic societal perspective and can introduce harm in downstream tasks \cite{bender2021dangers}. Making these models easier to control and use may encourage people to neglect the dangers involved with these models.

\bibliography{anthology,custom}
\bibliographystyle{acl_natbib}

\pagebreak 

\appendix


\section{Example Modified Probabilities}
\label{sec:logprobs}

Below is a figure that shows how our reweighting schemes adjust the log probability for a specific prompt.

\begin{figure}[h]
\centering
\includegraphics[width=.48\textwidth]{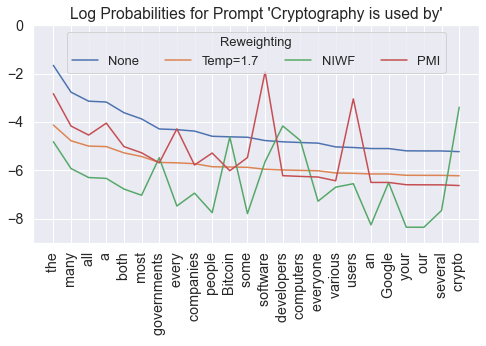}
\caption{NIWF and PPMI give more weight to more specific words like `Bitcoin' and `software'.}
\label{fig:logprobs}
\end{figure}

\section{Selecting First Noun Phrase}
\label{sec:nounphrase}

For the experiment, which is a prompt completion task, we generate 10 tokens and then parse the entire output (i.e. the prompt and the generated text) using Spacy.\footnote{https://spacy.io/} To select the first noun phrase, we choose either the first noun chunk, as tagged by Spacy, or the subtree of the first noun after the third generated word, whichever is longer.

\section{Diversity Measures}
\label{sec:diversity}

We follow \citet{takayama2020consistent} in their definitions of \textbf{dist} and \textbf{ent}. \textbf{Dist} \citep{li-etal-2016-diversity} is defined as the number of distinct $n$-grams in the generated outputs divided by the total number of generated tokens. \textbf{Ent} \citep{NEURIPS2018_23ce1851} considers the frequency of $n$-grams in the generated outputs, such that

\begin{equation*}
    \text{ent} = -\frac{1}{\sum_w F(w)} \sum_{w \in Y} F(w) \log \frac{F(w)}{\sum_w F(w)}
\end{equation*}

where $Y$ is a set of $n$-grams output by the system and $F(w)$ is the frequency of each $n$-gram. 
We look at all generated responses per topic in a given condition to calculate the diversity measures, and then average the measures across the five topics.
We report both dist-1 (unigrams) and dist-2 (bigrams) as well as ent-2 (bigrams). 


\end{document}